\documentclass[10pt,twocolumn,letterpaper]{article}

\usepackage{cvpr}              

\usepackage[dvipsnames]{xcolor}

\definecolor{cvprblue}{rgb}{0.21,0.49,0.74}
\usepackage[pagebackref,breaklinks,colorlinks,citecolor=cvprblue]{hyperref}
\usepackage{tabularx}
\usepackage{algorithm}
\usepackage{algpseudocode}

\def\paperID{3806} 
\def\confName{CVPR}
\def\confYear{2024}

\usepackage{amsmath,amsfonts,bm}

\def\eqref#1{equation~\ref{#1}}

\def\1{\bm{1}}

\def\rvc{{\mathbf{c}}}

\def\rvv{{\mathbf{v}}}

\def\rmK{{\mathbf{K}}}

\def\rmR{{\mathbf{R}}}

\DeclareMathAlphabet{\mathsfit}{\encodingdefault}{\sfdefault}{m}{sl}
\SetMathAlphabet{\mathsfit}{bold}{\encodingdefault}{\sfdefault}{bx}{n}

\def\gC{{\mathcal{C}}}

\def\gE{{\mathcal{E}}}

\def\gI{{\mathcal{I}}}

\def\gL{{\mathcal{L}}}
\def\gM{{\mathcal{M}}}

\def\gO{{\mathcal{O}}}
\def\gP{{\mathcal{P}}}

\def\gV{{\mathcal{V}}}
\def\gW{{\mathcal{W}}}

\newcommand{\redtext}[1]{\textcolor{red}{#1}}

\newcommand\mypara[1]{\vspace{1mm}\noindent\textbf{#1.}}

\title{L-MAGIC: \redtext{L}anguage \redtext{M}odel \redtext{A}ssisted \redtext{G}eneration of \redtext{I}mages with \redtext{C}oherence}

\author{Zhipeng Cai\thanks{Corresponding author (zhipeng.cai@intel.com)}\ \ \ \  Matthias Mueller\ \ \ \  Reiner Birkl \ \ \ \  Diana Wofk \ \ \ \  Shao-Yen Tseng \\  
Junda Cheng \ \ \ \ Gabriela Ben-Melech Stan \ \ \ \  Vasudev Lai \ \ \ \ Michael Paulitsch\\
Intel Labs
}

\begin{document}
\maketitle
\begin{abstract}
In the current era of generative AI breakthroughs, generating panoramic scenes from a single input image remains a key challenge. Most existing methods use diffusion-based iterative or simultaneous multi-view inpainting. However, the lack of global scene layout priors leads to subpar outputs with duplicated objects (\eg, multiple beds in a bedroom) or requires time-consuming human text inputs for each view. We propose L-MAGIC, a novel method leveraging large language models for guidance while diffusing multiple coherent views of $360^\circ$ panoramic scenes. L-MAGIC harnesses pre-trained diffusion and language models without fine-tuning, ensuring zero-shot performance. The output quality is further enhanced by super-resolution and multi-view fusion techniques. Extensive experiments demonstrate that the resulting panoramic scenes feature better scene layouts and perspective view rendering quality compared to related works, with $>$$70\%$ preference in human evaluations.
Combined with conditional diffusion models, L-MAGIC can accept various input modalities, including but not limited to text, depth maps, sketches, and colored scripts. Applying depth estimation further enables 3D point cloud generation and dynamic scene exploration with fluid camera motion. Code is available at \url{https://github.com/IntelLabs/MMPano}. 
\end{abstract}
\section{Introduction}
\label{sec:intro}

\begin{figure}
    \centering
    \includegraphics[clip, trim=0.9cm 5.2cm 0.5cm 0.3cm, width=.8\columnwidth]{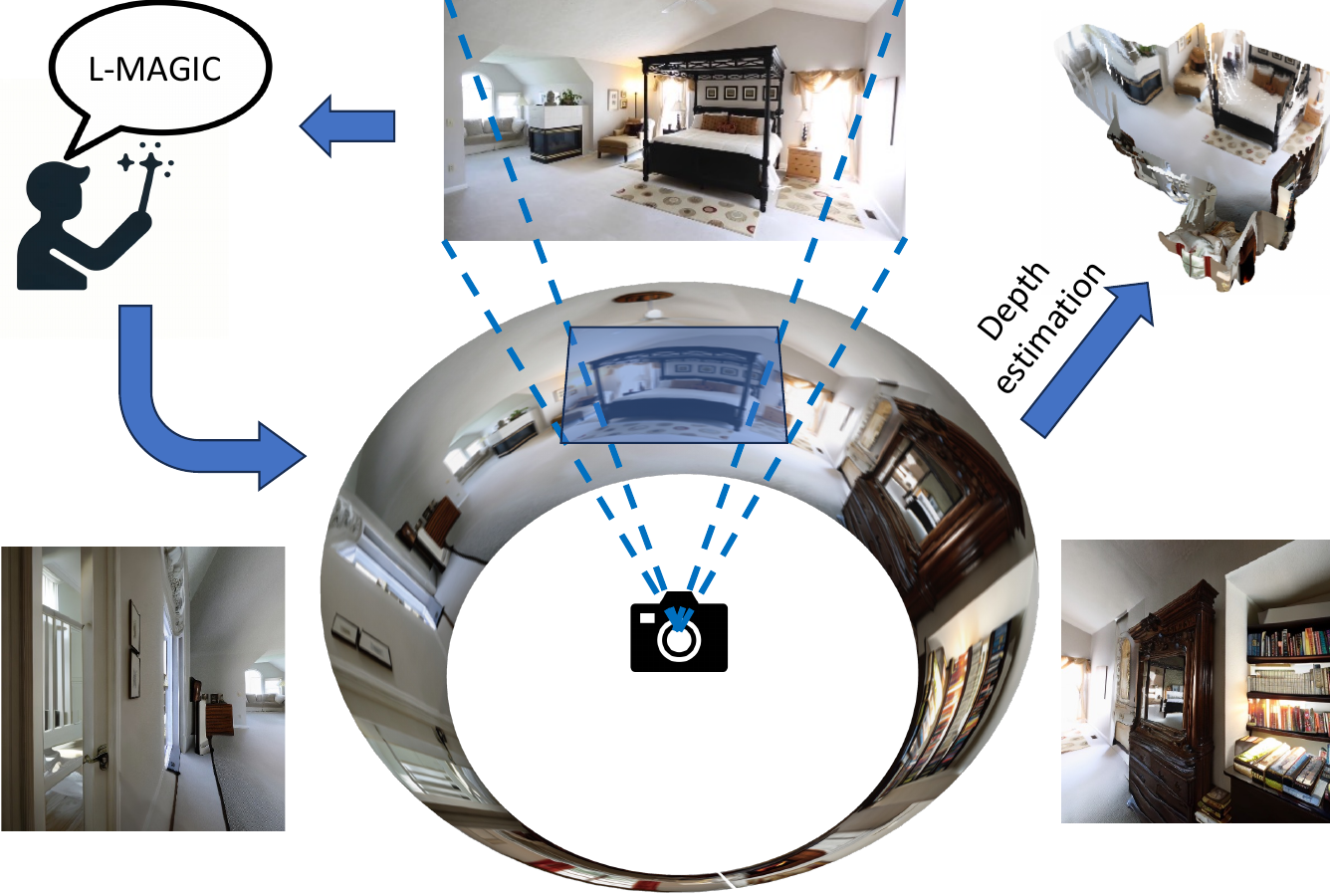}
    \caption{\textbf{Teaser.} L-MAGIC is a novel method to generate a $360^\circ$ panoramic scene from a single input image. L-MAGIC utilizes large language models to control perspective diffusion models to generate multiple views with coherent $360^\circ$ layout. L-MAGIC is also compatible with images synthesized by conditional generative models, making it capable of creating panoramic scenes from various input modalities. A set of perspective images rather than a single panoramic image also allows our method to leverage off-the-shelf monocular depth estimation models to enable immersive experiences, \eg, scene fly-through or 3D point cloud generation.}
    \vspace{-6pt}
    \label{fig:teaser_intro}
\end{figure}


Diffusion models have achieved state-of-the-art performance in image generation. However, generating a $360^\circ$ panoramic scene from a single perspective image remains a challenge, which is an important problem in many computer vision applications, such as architecture design, movie scene creation, and virtual reality (VR). 

Training a model to directly generate panoramic images is challenging due to the lack of diverse large-scale datasets. Hence, most existing works separate panoramic scenes into multiple perspective views, and inpaint them using pre-trained diffusion models. To ensure generalization, the diffusion model is either frozen without any architecture change~\cite{text2room} or combined with extra modules trained on small datasets for integrating multi-view information~\cite{MVDiffusion}. 

A common approach to encode the scene information during multi-view inpainting is to provide a text-conditioned diffusion model with the description of the input image, which is generated by a user or an image captioning model~\cite{BLIPv2}. Though effective for extending local scene content, such approaches suffer from \emph{incoherence} in the overall scene layout. Specifically, using the same text description for diffusing different views along the $360^\circ$ panorama leads to artifacts and unnecessarily repeated objects. Current inpainting methods have no mechanism to leverage global scene information in individual views. 


In this work, we show that state-of-the-art (vision) language models, without fine-tuning, can be used to control multi-view diffusion and effectively address the above problem. We propose {L-MAGIC} (Fig.~\ref{fig:teaser_intro}), a novel framework leveraging large language models to enable the automatic generation of diverse yet coherent $360^\circ$ views from a given input image. L-MAGIC relies on iterative warping-and-inpainting. Pre-trained (vision) language models are used to: (1) generate layout descriptions of different views that are used in text-conditioned inpainting, (2) automatically determine whether salient objects should be repeated or not for a specific scene, and (3) monitor the inpainting outputs to avoid challenging cases where diffusion models violate the text guidance. A key contribution is the prompt design for language and diffusion models to make L-MAGIC \emph{fully automatic}. In addition, smooth multi-view fusion and super-resolution techniques are used to ensure high resolution and quality when producing the final panorama. 


Experiments show that L-MAGIC generates $360^\circ$ panoramic scenes with higher quality and more coherent layouts compared to state-of-the-art methods. Not relying on model fine-tuning makes L-MAGIC effective on in-the-wild images, and extendable, using conditional diffusion models~\cite{stable_diffusion, controlnet}, to other types of inputs such as text, depth maps, sketch drawings and color scripts/segmentation masks. Applying depth estimation further enables the creation of 3D point clouds and immersive scene fly-through experiences with both camera rotation and translation. 

\section{Related Work}
\label{sec:related_work}

\mypara{Diffusion models} Diffusion models learn to generate data by inverting the diffusion process, \ie, removing the noise in the data (see~\cite{yang2022diffusion} for a detailed survey). By separating the data generation process into multiple noise removal steps (reverse process)~\cite{ho2020denoising}, diffusion models learn image synthesis much more effectively than GANs~\cite{GAN}. Recently, latent diffusion models~\cite{stable_diffusion} have been proposed to improve the training speed and image synthesis quality by performing the diffusion in latent space. By training the model on large-scale image-caption pairs~\cite{laion}, they achieve remarkable quality and robustness in text-conditioned image synthesis. Further fine-tuning of latent diffusion models using large mask strategies~\cite{lama} produces robust text-conditioned inpainting models, which are used in this work.

\mypara{Panoramic scene generation} Various approaches have been proposed for panoramic scene generation. Some of them~\cite{text2light, LDM3D} treat the panorama as a single equirectangular image, and generate it in a single forward pass. However, such approaches struggle to close the loop at both ends of the generated equirectangular image, even with purposely designed spherical positional embeddings~\cite{text2light}. Meanwhile, the lack of large scale training data makes it impractical to train a generalizable model for image-to-panorama, and limits the robustness of the model, resulting in inconsistent outputs with the input descriptions. 

Some recent methods~\cite{text2room, MVDiffusion} create panoramas by generating multiple perspective views using robust pre-trained diffusion models trained on large-scale perspective data. Text2room~\cite{text2room} generates 10 rotated views of a panorama using stable diffusion v2 inpainting without layout control and focuses on indoor scenes and mesh generation. MVDiffusion~\cite{MVDiffusion} ensures multi-view local texture consistency by extra multi-view attention modules fine-tuned on small datasets. Though applicable to both text-to-panorama and image-to-panorama, these methods struggle to generate diverse $360^\circ$ views. Specifically, there is no mechanism to encode the global scene layout into the generation or inpainting of different views. Hence, conditioning the method only on the input image or text results in salient objects (\eg, beds in a bedroom) being generated repeatedly across views. In this work, we guide the multi-view diffusion process with large language models to automatically generate panoramic scenes with coherent and diverse $360^\circ$ layouts.

\mypara{Language models} Recent language model advancements have enabled many important applications (see~\cite{zhao2023survey} for details). Trained on large-scale data with humans in the loop, ChatGPT~\cite{chatGPT} has demonstrated super-human performance on various language-based tasks. In this work, we utilize ChatGPT to automatically generate coherent multi-view scene descriptions for the consumption of a pre-trained diffusion model that generates multiple perspective views of a panoramic scene. Leveraging multi-modal data, vision language models~\cite{zhang2023vision} have further enhanced language models to understand visual inputs. In this work, we utilize pre-trained VQA models~\cite{BLIPv2} to automatically generate a scene description for the input image, and to avoid unnecessarily repeated objects across the generated panoramic scene.
\begin{figure*}
   \centering
   \includegraphics[width=.9\textwidth]{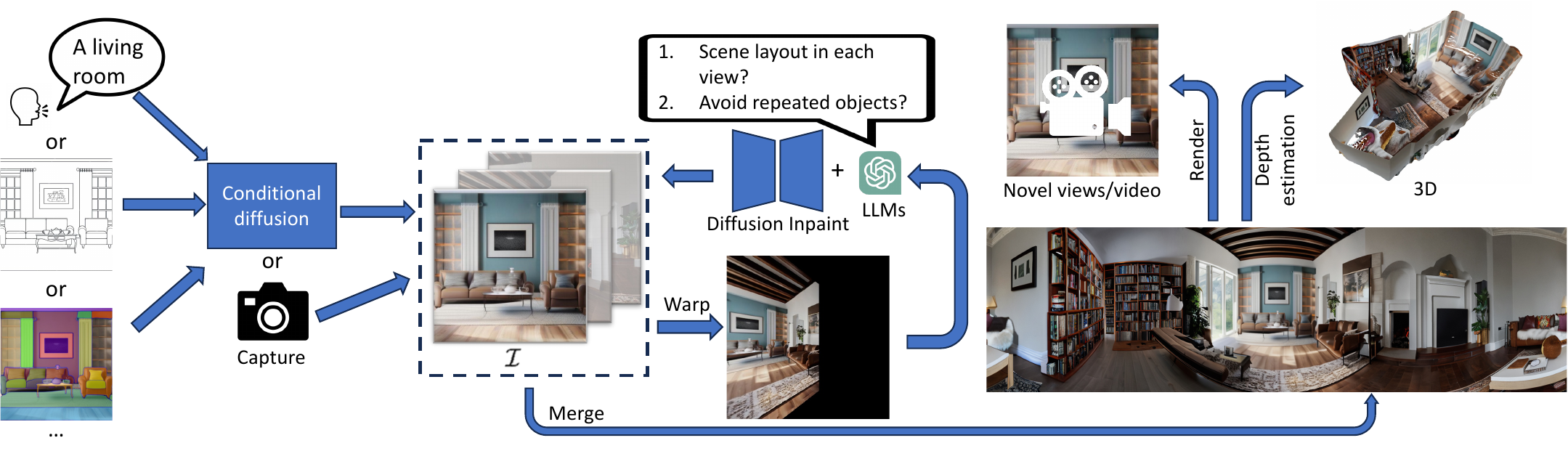}
   \caption{\textbf{L-MAGIC pipeline.} The input is an image $\gI$ either captured in the real-world or synthesized, \eg, by conditional diffusion models. Multiple novel views to compose a $360^\circ$ panoramic scene are generated by iterative warping and inpainting. Pre-trained diffusion models assisted by pre-trained language models are used to generate views with both high-quality local textures and coherent $360^\circ$ layouts. Further quality enhancement techniques ensure smooth blending of multiple views into high-resolution panoramic scenes. L-MAGIC can generate panorama images, immersive videos, and 3D point clouds from various types of inputs, such as images, text, and sketch drawings.}
   \label{fig:pipeline}
   \vspace{-1em}
\end{figure*}

\section{Methdology}\label{sec:method}

The goal of this work is to generate a coherent $360^\circ$ panoramic scene given a single (perspective) image. Note that this setting is very general since the input image could be either captured in the real world or synthesized. For example using conditional diffusion models~\cite{stable_diffusion, controlnet}, one can synthesize the input image using inputs such as text descriptions, sketches, depth images, and so on. 



As shown in Fig.~\ref{fig:pipeline}, L-MAGIC generates the panoramic scene with iterative warping-and-inpainting. The \emph{warping} step generates an incomplete perspective view and the mask of the missing region (Sec.~\ref{sec:warp}). The \emph{inpainting} step completes the masked region with assistance from language models (Sec.~\ref{sec:inpaint}). The final panorama is created by fusing the generated views with some post-processing to enhance the quality and resolution (Sec.~\ref{sec:method_quality}).

\subsection{Warping}\label{sec:warp}

At each warping step, we project all completed views onto a unit sphere representing the panoramic scene, and then render the next incomplete view to inpaint based on the relative camera pose. To project an image to the unit sphere, we first construct a mesh by defining the vertices $\gV$ on each image pixel and creating edges $\gE$ between adjacent pixels. Then, we project the vertices to a unit sphere by 
\begin{equation}\label{eq:map_sphere}
    \rvv_\text{sp} = \frac{\rmK^{-1} \rvv}{||\rmK^{-1} \rvv||},
\end{equation}
where $\rmK$ is the intrinsic matrix, $\rvv \in \gV$ is the homogeneous coordinate~\cite{sturm2005multi} of a pixel, and $\rvv_\text{sp}$ is the projected location. To warp a completed view $A$ to a novel view $B$, where $\rmR$ is the rotation matrix from $A$ to $B$, we rotate each projected vertex $\rvv_\text{sp}$ of $A$ by $\rvv_\text{rot} = \rmR\rvv_\text{sp}$, and then perform rasterization-based rendering~\cite{pytorch3d} $(\gI, \gM) = \text{rasterize}(\gV_\text{rot}, \gE, \rmK')$ where $\gV_\text{rot}$ is the set of rotated vertices and $\rmK'$ is the intrinsic matrix of image $B$. The output $\gI$ is a warped image and $\gM$ is a binary mask indicating whether inpainting is required for a pixel (obtained by checking for each pixel whether ray-casting hits a valid mesh face).

To ensure the local inpainting consistency of each perspective view, we use a large field of view (FoV) and adjust the rotation angles so that both known and unknown regions are reasonably large after warping. In practice, a FoV of 100 degrees with roughly 40 degrees of rotation between adjacent views works well. To further reduce the iterative error accumulation, we expand the scene alternatively from both sides of the input image, rather than expanding in a single direction. To ensure a smooth $360^\circ$ loop closure, we tune the rotation angles so that the final view has a large incomplete region at the center, resulting in a sequence of views with rotation angles of $\{0^\circ, 41^\circ, -41^\circ, 82^\circ, -82^\circ, 123^\circ, 200.5^\circ \text{(for loop closure)}\}$. 

\subsection{Inpainting with Language Model Assistance}\label{sec:inpaint}

The inpainting step completes a warped view with a consistent local style and a coherent $360^\circ$ scene layout. 
We utilize the Stable Diffusion v2 inpainting model~\cite{stable_diffusion}, which can effectively extrapolate the large missing region of each warped view while maintaining local style consistency. However, naive inpainting without any prior will generate severe artifacts. One common prior explored before~\cite{text2room, MVDiffusion} is a user-provided text description of the scene or input image.
Yet, using the same description in different views may generate duplicate objects such as multiple beds in a bedroom (see \cref{fig:qual_i2p}), since perspective inpainting methods have no mechanism to split the layout into different views. Please refer to \cref{sec:exp_analysis} for a detailed ablation study.

\begin{algorithm}[t] 
\caption{L-MAGIC}
\label{alg:main}
\begin{algorithmic}[1]
    \State \textbf{Input}: Initial image $\gI$, intrinsic matrix $\rmK$ of $\gI$, intrinsics $\rmK'$ and poses $\gP = \{\rmR_i\}_{i=1}^N$ of warped views, vision language model $\gL_\text{v}(\cdot)$, language model $\gL(\cdot)$, text-conditioned inpainting model $f_\text{inpaint}(\cdot)$.
    \State $d_{\gI} \leftarrow \gL_\text{v}(\gI, \text{`Description of input image'})$ \label{line:d_init}
    \State $d_{360} \leftarrow \gL(d_{\gI}, \text{`Layout of individual views'})$ \label{line:d_360}
    \State $d_{\text{scene}} \leftarrow \gL(d_{\gI}, \text{`Remove object-level descriptions'})$ \label{line:d_scene}
    \State $d_\text{repeat} \leftarrow \gL(d_{\gI}, \text{`Avoid duplicated objects'})$ \label{line:d_repeat}
    \State $\gW_1, \gM_1 \leftarrow \text{warp}(\gC, \rmK, \rmK', \rmR_1)$ \label{line:warp_init}
    \State $\gI_1 \leftarrow f_\text{inpaint}(\gW_1, \gM_1, d_{\text{scene}})$ \Comment{expand the FoV of $\gI$} \label{line:inpaint_init}
    \State $\gC \leftarrow \{\gI_1\}$
    \For{$i=2$ \textbf{to} $N$}
      \State $c \leftarrow 0$ \label{line:count}
      \State $\gW_i, \gM_i \leftarrow \text{warp}(\gC, \rmK, \rmK', \rmR_i)$ \label{line:warp}
      \State $d_i \leftarrow \text{generate$\_$prompt}(d_{360}, d_\text{scene}, d_\text{repeat}, i)$ \label{line:prompt}
      \State $\gI_i \leftarrow f_\text{inpaint}(\gW_i, \gM_i, d_i)$ \label{line:inpaint}
      \For{each object $\gO$ in $d_\text{repeat}$}
        \If{$\gL_\text{v}(\gI_i, \text{`Is $\gO$ in $\gI_i$?'}) = \text{'yes'} \bigcap c < 20$} \label{line:check}
            \State $c \leftarrow c + 1$
            \State Go to line~\ref{line:inpaint}.\label{line:re-inpaint}
        \EndIf
      \EndFor
      \State $\gC \leftarrow \gC \bigcup \{\gI_i\}$ \label{line:add_view}
    \EndFor
    \State \Return merge($\gC$) \label{line:return}
\end{algorithmic}
\end{algorithm}

To address these problems, we use a vision language model $\gL_\text{v}(\cdot)$ (BLIP-2~\cite{BLIPv2}) and a language model $\gL(\cdot)$ (ChatGPT4~\cite{chatGPT}) to guide the inpainting process. 
In the following, we describe our method (Alg.~\ref{alg:main}) in detail. The exact prompts used in Alg.~\ref{alg:main} to interact with language and diffusion models are provided in Appendix~\ref{appdx:questions}.

Before warping and inpainting, we first prompt $\gL_\text{v}(\cdot)$ to generate the description $d_\gI$ for the input image $\gI$ (line~\ref{line:d_init}).  
We ask two questions so that $d_\gI$ contains both coarse and fine levels of detail.
Next, we ask $\gL(\cdot)$ to imagine the global scene layout $d_{360}$ (line~\ref{line:d_360}) based on $d_\gI$, where each line of $d_{360}$ corresponds to the description of a specific view. To avoid duplicate objects, we request a compact description of individual views without mentioning objects in other views. 

$d_{360}$ mostly contains objects of individual views. Using such descriptions as the inpainting prompt can lead to inconsistent style at distant views. Hence, we ask $\gL(\cdot)$ to remove objects from $d_\gI$ and obtain the final scene-level description $d_\text{scene}$, \eg, `a bedroom with a wooden bed' becomes `a bedroom' (line~\ref{line:d_scene}). $d_\text{scene}$ is later used together with $d_{360}$ to ensure a consistent multi-view style. 

Though $d_\text{scene}$ ensures the multi-view style consistency, the training data bias of diffusion models may still result in objects commonly associated with a particular scene being generated, even if not explicitly mentioned in $d_{360}$. For example, a bed is often generated with the word `bedroom' in the prompt, resulting in duplicate beds in multiple views. To avoid this problem, we further let $\gL(\cdot)$ automatically determine whether there are some objects in the scene that require repetition avoidance (line~\ref{line:d_repeat}).

After each warping step, we use the outputs from lines~\ref{line:d_init} to~\ref{line:d_repeat} to automatically generate the prompt for text-conditioned inpainting (line~\ref{line:prompt}). Specifically, for the warped view with $0^\circ$ rotation (i = 1), we use $d_\text{scene}$ as the prompt ($d_i$) for text-conditioned inpainting (line~\ref{line:inpaint_init}). For other views (line~\ref{line:inpaint}), if there is no object in $d_\text{repeat}$, \ie, no repetition avoidance required, we perform inpainting with the prompt \emph{`a peripheral view of $<$$d_\text{scene}$$>$ where we see $<$the corresponding description in $d_{360}$$>$'}. If any object exists in $d_\text{repeat}$, we use the positive prompt of \emph{`a peripheral view of $<$$d_\text{scene}$$>$ where we only see $<$the corresponding description in $d_{360}$$>$'}, and the negative prompt of \emph{`any type of $<$the object in $d_\text{repeat}$$>$'} (one sentence for each object in $d_\text{repeat}$). The positive prompt template prevents Stable Diffusion from generating common objects of an environment (\eg, the bed in a bedroom). The negative prompt template avoids duplication of objects mentioned in $d_\text{repeat}$. 

Bias exists in the training data of diffusion models --- an image with the caption of `a bedroom' mostly contains a bed. Therefore, repeated objects can still be generated even with constraints from the prompt. To further alleviate this problem, we use $\gL_\text{v}(\cdot)$ to check whether each inpainted image $\gI_i$ contains objects mentioned in $d_\text{repeat}$ (line~\ref{line:check}).
If the answer is `yes', we re-run inpainting until the answer becomes `no' or the maximum number of trials $c$ is reached.


\subsection{Quality and Resolution Enhancement}\label{sec:method_quality}
Several techniques are also proposed to enhance the quality and resolution of the final panorama.


As shown in Appendix.~\ref{appdx:blur}, adjacent pixels at the center of an image have a larger angular distance than the ones at the side of an image. When warping a completed view to a novel view, the original central region becomes the side region, making the rendered image blurry due to interpolation. Meanwhile, the resolution of the Stable Diffusion output is $512 * 512$, the panorama created by these images has a low resolution. To address both problems, we apply super-resolution~\cite{SD4x} to the output $\gI_i$ of each inpainting step, increasing the resolution of $\gI_i$ to $2048*2048$. Then, we warp the high-resolution image to a low-resolution novel view so that no (strong) interpolation is required. After performing all warping and inpainting steps, we simply fuse the super-resolution images to generate a high-resolution panorama.


During warping and panorama generation, multiple perspective images might have overlaps at the same region. To avoid sharp boundaries when merging them, we perform a weighted average, \ie, given multiple warped pixels at the same location with colors $\rvc_i$, the final merged pixel color is $\rvc_\text{merge} = \frac{\sum_i w_i \rvc_i}{\sum_i w_i}$, where the weight $w_i$ is computed as the distance to the nearest image boundary at the original view $i$. This strategy effectively down-weights the pixels near the warping boundaries, ensuring a smooth transition during multi-view fusion.

To create the final panorama (line~\ref{line:return}), we first project each view to the unit sphere same as in Sec.~\ref{sec:warp}. Then, we perform the equirectangular projection~\cite{equi_proj} to warp multiple projected views to the same equirectangular plane, and merge them into a single equirectangular image.

\subsection{Discussion}\label{sec:discussion}

L-MAGIC is \emph{fully automatic} --- no human interaction is required to link language models and diffusion models. This is realized by 1) careful prompt engineering, which enables language models to output texts that can be automatically converted into the inpainting prompt, and 2) handling the edge cases where language models or diffusion models do not generate outputs that satisfy the requirements in the input prompt. For example, ChatGPT sometimes still outputs the layout description $d_{360}$ with an erroneous format, making automatic prompt extraction fail catastrophically. We automatically detect such cases, and re-run line~\ref{line:d_360} to ensure the algorithm flow, see Appendix~\ref{appdx:questions} for more details. 

L-MAGIC requires \emph{no model fine-tuning}, which ensures the zero-shot performance and makes it capable of accepting other types of inputs leveraging conditional generative models (see Sec.~\ref{sec:exp} for results). This advantage also allows individual modules to be replaced by future methods to enhance the performance, \eg, change BLIP-2+ChatGPT to GPT-4V~\cite{GPT-4V}, or use other inpainting models. 



\section{Experiments}\label{sec:exp}

We describe our experimental setup in \cref{sec:exp_setup} and compare our method with other panorama generation methods in \cref{sec:exp_main}. We then analyze the contribution of individual components of our methods in \cref{sec:exp_analysis}. Finally, we demonstrate several down-stream applications, such as scene fly-through and 3D scene generation in \cref{sec:exp_application}. 

\subsection{Experimental Setup}\label{sec:exp_setup}

\mypara{Baselines} We evaluate our method on both image-to-panorama and text-to-panorama. For \emph{image-to-panorama}, we compare against:
\begin{enumerate}
    \item Stable Diffusion v2~\cite{stable_diffusion}: we use the prompt `360 degree panorama of $<$scene description$>$' to inpaint the panorama image in a single diffusion process.
    \item Text2room~\cite{text2room}: we take the panorama generation component (at steps 11-20 of the pipeline) for comparison. 
    \item MVDiffusion image-to-panorama model~\cite{MVDiffusion}. 
\end{enumerate}
\vspace{3pt}
To enable text conditioning in these methods, we use BLIP-2 to obtain the description of the input image. For \emph{text-to-panorama}, we compare against:
\begin{enumerate}
    \item Text2light~\cite{text2light}: GAN-based text to panorama model. 
    \item Stable Diffusion v2~\cite{stable_diffusion}: we use the prompt `360 degree panorama of $<$input prompt$>$' to generate the panoramic image in a single diffusion process.
    \item LDM3D~\cite{LDM3D} panorama model: we only use the output RGB image and the prompt `360 degree panorama of $<$input prompt$>$' to generate the panoramic image in a single diffusion process.
    \item Text2room panorama generation module ~\cite{text2room}. 
    \item MVDiffusion text-to-panorama model~\cite{MVDiffusion}. 
\end{enumerate}

\mypara{Implementation Details} L-MAGIC is implemented with PyTorch~\cite{pytorch} and the official release of BLIP-2~\cite{BLIPv2}, Stable Diffusion~\cite{stable_diffusion} and the ChatGPT4 API~\cite{chatGPT}. It takes 2-5 minutes to generate a $360^\circ$ panorama depending on the number of repetitions in line~\ref{line:re-inpaint} of Alg.~\ref{alg:main}. We take the official code and model for all other methods. In text-to-panorama, L-MAGIC uses Stable Diffusion v2 conditioned on the given text prompt to generate the input image. 


\mypara{Data} To evaluate the in-the-wild performance, we collect data that does not overlap with the training data of any method for both tasks. For image-to-panorama, we use 20 indoor and 20 outdoor images from tanks-and-temples~\cite{TaT} and RealEstate10K~\cite{RealEstate10K}. For text-to-panorama, we use ChatGPT to generate 20 random scene descriptions (10 indoor and 10 outdoor, see Appendix~\ref{appdx:text2pano}). 

\mypara{Metrics} To evaluate different methods with respect to quality and multi-view diversity, we compute the Inception Score (IS)~\cite{inception} for the perspective views of the panorama. Since existing quantitative metrics do not capture all aspects of human perception of quality~\cite{barratt2018note}, we follow existing works~\cite{text2room, text2light, repaint} for a complementary human evaluation. To this end, we set up a voting web page that shows side-by-side panoramic scenes generated using the same input, one with our method and one with a baseline. We ask 15 anonymous voters to choose which panorama has higher quality and scene structure (see Appendix~\ref{appdx:eval_webpage} for an example voting page). To minimize voting bias, we randomly shuffle the order of the side-by-side panoramas and hide the generation method names. We use the votes to compute a preference score from 0 to 1 for our method compared to the baselines. This score is simply the percentage of votes for our method with respect to quality and structure.

\subsection{Main Results}\label{sec:exp_main}

\begin{figure*}[t]
    \centering
    \includegraphics[width=1.\textwidth]{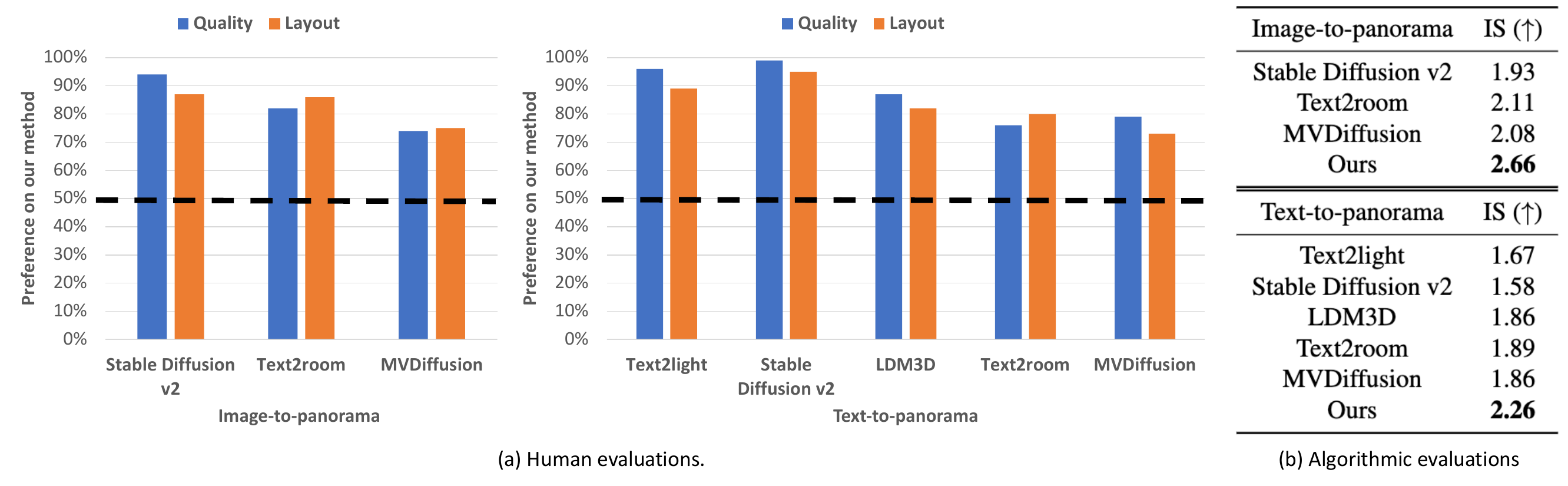}
    \vspace{-2em}
    \caption{\textbf{Quantitative results for image-to-panorama and text-to-panorama.} (a) Human evaluations. Each baseline has two bars representing respectively the quality of rendered perspective views and the $360^\circ$ layout. The value of the bar means the frequency where our method is preferred in the voting. Above $50\%$ (dashed line) means our method is more preferred than the corresponding baseline. (b) Algorithmic evaluation by computing the Inception Score (IS). L-MAGIC consistently outperforms previous methods on both metrics.}
    \label{fig:exp_main}
\end{figure*}

\begin{figure*}[t]
    \centering
    \includegraphics[width=1.\textwidth]{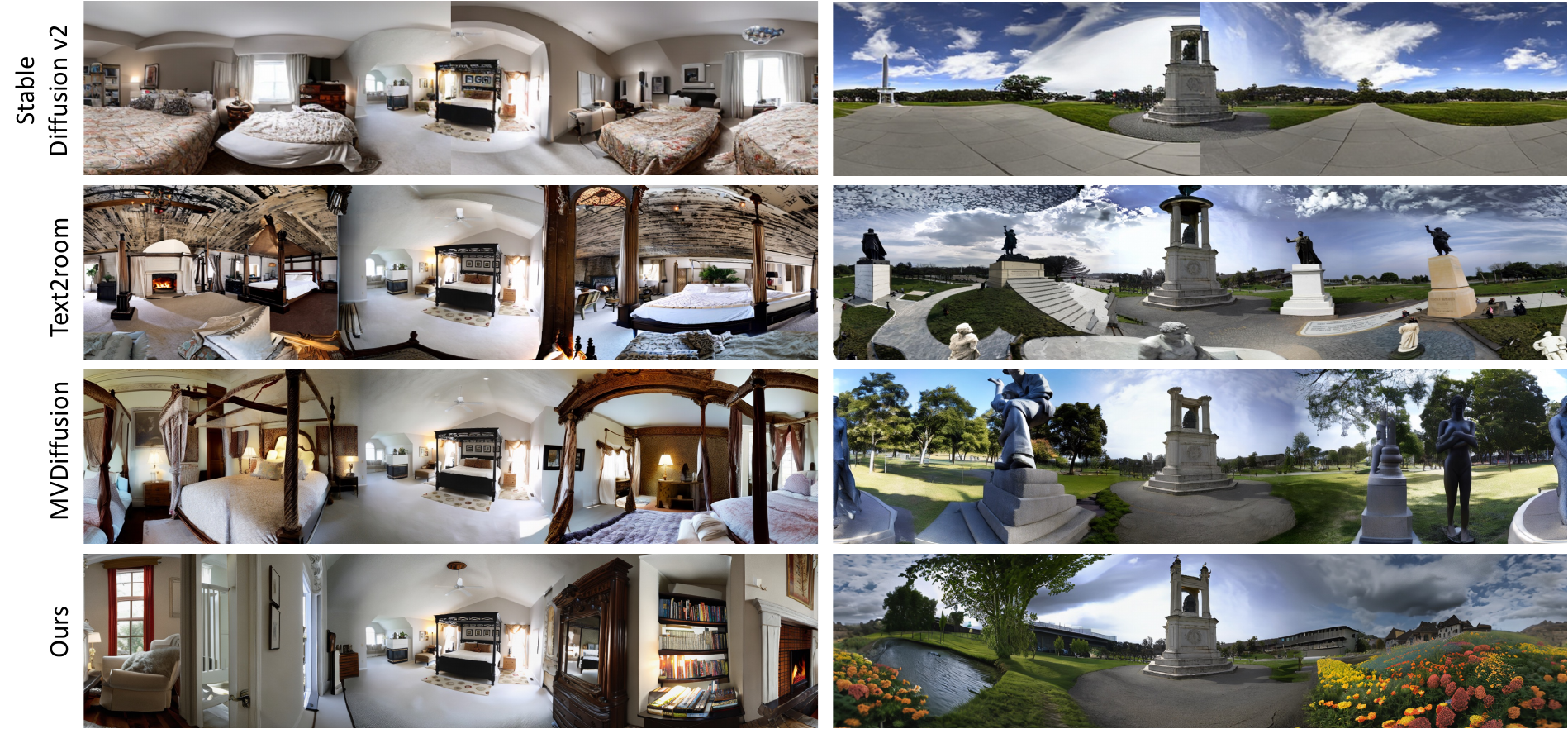}
    \vspace{-1em}
    \caption{\textbf{Image-to-panorama visualizations.} Stable Diffusion v2 cannot close the $360^\circ$ loop (sharp boundaries at the middle). Text2room and MVDiffusion lack mechanisms to avoid duplicate objects. L-MAGIC outputs have high local view quality and coherent scene layouts.}
    \label{fig:qual_i2p}
\end{figure*}

\begin{figure*}[t]
    \centering
    \includegraphics[width=1.\textwidth]{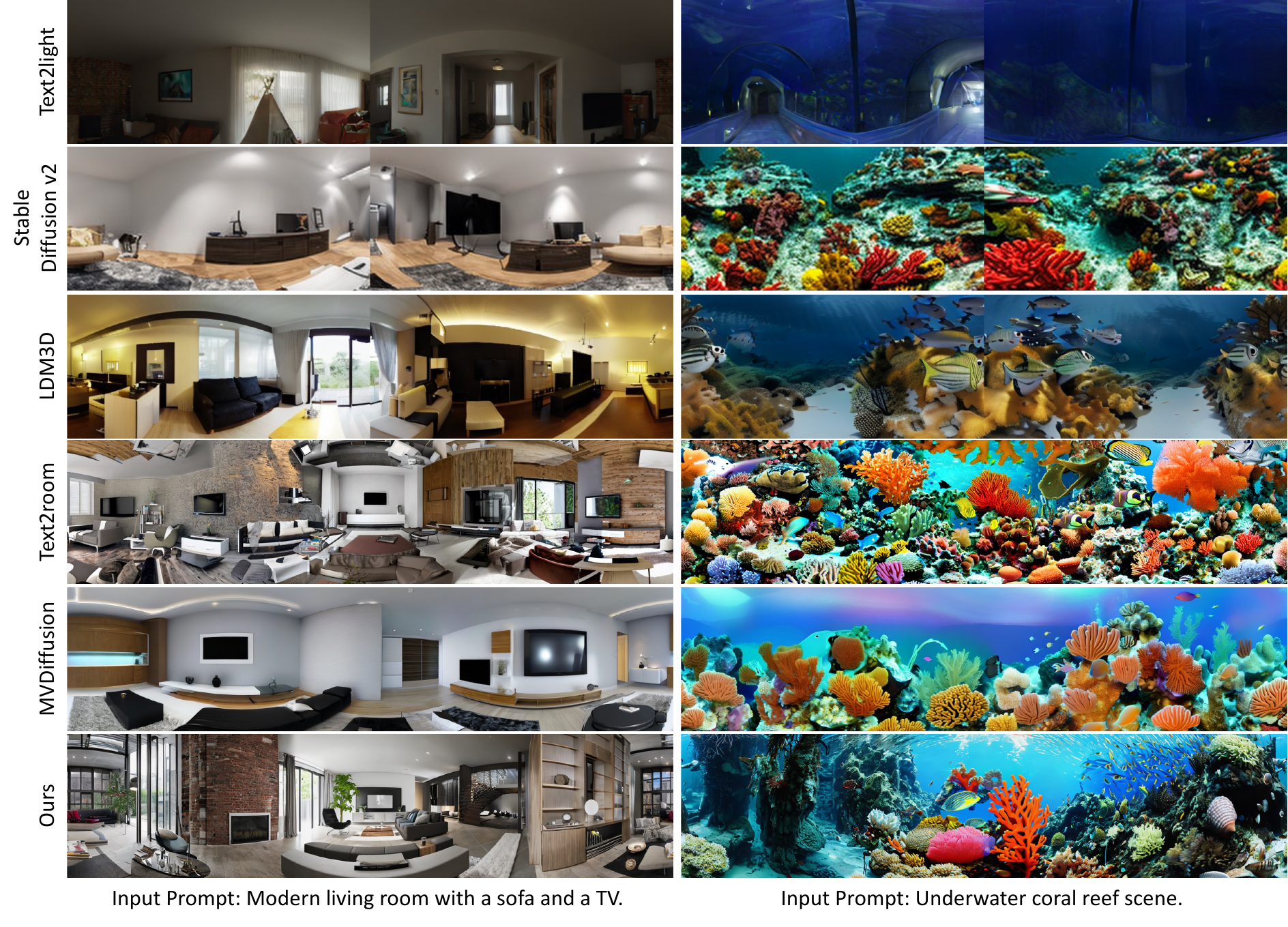}
    \vspace{-1.5em}
    \caption{\textbf{Text-to-panorama visualizations.} Text2light, Stable Diffusion v2 and LDM3D cannot close the $360^\circ$ loop (sharp boundaries at the middle). Text2room and MVDiffusion generate panoramas with duplicate objects. L-MAGIC effectively addresses these problems, resulting in high-quality panorama with reasonable scene layouts.}
    \label{fig:qual_t2p}
\end{figure*}

As shown in Fig.~\ref{fig:exp_main}, our method performs better for both image-to-panorama and text-to-panorama, even compared to task-specific methods. To further understand the performance of different methods, we provide in Fig.~\ref{fig:qual_i2p} and~\ref{fig:qual_t2p} the qualitative results for both tasks. \emph{Stable Diffusion v2} treats a panorama as a single image. It cannot close the $360^\circ$ loop since equirectangular projection~\cite{equi_proj} splits the loop-closure area to two sides of an image (moved to the middle of the rendered panorama for better visualization). In the meantime, due to the lack of large-scale panorama training data, it still generates unnecessarily repeated objects such as multiple beds in a bedroom. \emph{Text2room} and \emph{MVDiffusion} separate a panorama into multiple perspective views. Inpainting them using the same prompt results in unreasonably repeated objects in multiple views. Due to the limited panorama training data, \emph{Text2light} cannot fully understand zero-shot scene descriptions generated by ChatGPT, resulting in scenes not consistent with the input prompt. Similar to Stable Diffusion v2, treating panorama as a single image makes it fail on loop closure. \emph{LDM3D} is fine-tuned on top of a perspective latent diffusion model. Though better than Text2light, it still cannot close the loop and sometimes fails to generate scenes that are consistent with the details of the prompt (\eg, generating a non-modern living room when asked for a modern one). \emph{Our method} works robustly on various inputs, generating panoramic scenes with high perspective rendering quality and reasonable $360^\circ$ scene layouts (see supplementary videos for more details).






\subsection{Analysis}\label{sec:exp_analysis}

\begin{figure}[t]
    \centering
    \includegraphics[width=1\columnwidth]{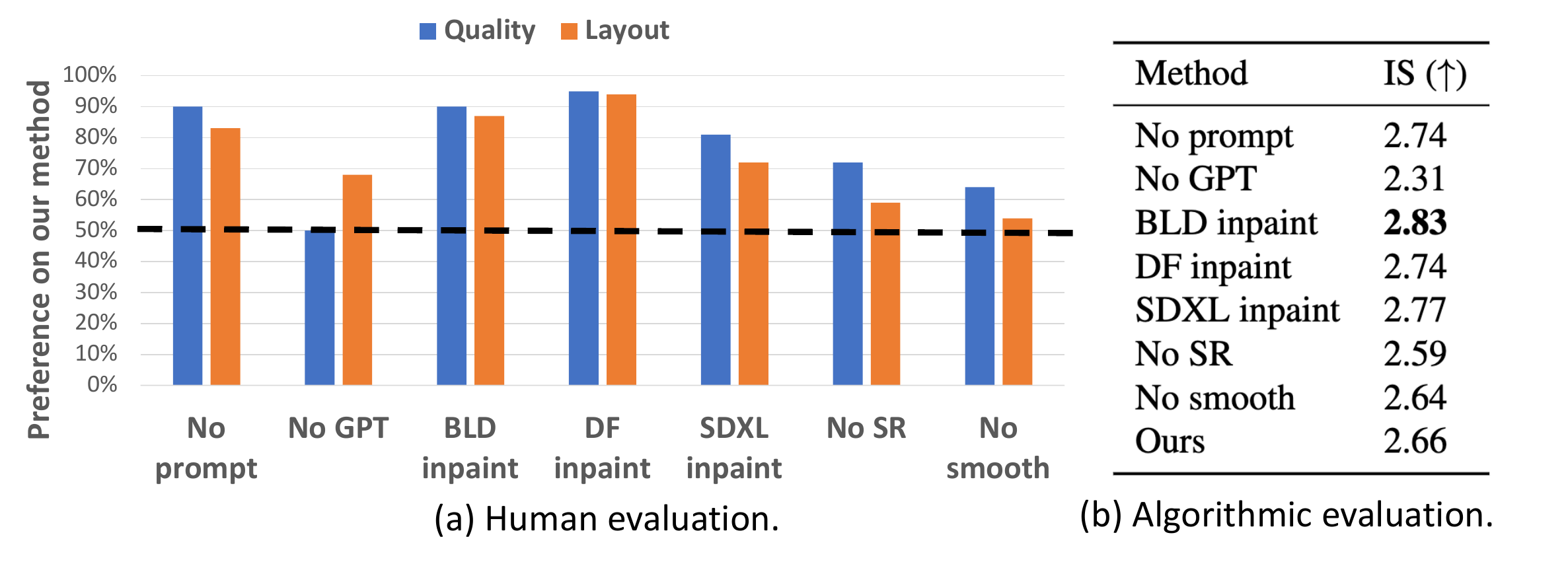}
    \caption{\textbf{Analysis.} We follow the same evaluation protocol as in the main experiment (Fig.~\ref{fig:exp_main}). From human evaluations, we see that removing the language assistance results in severe degration in either the scene layout (No GPT) or both the quality and the layout (No Prompt). Replacing Stable Diffusion v2 with other inpainting methods results in performance drop on both the quality and the layout. The rendering quality decreased with quality enhancement techniques removed. The Inception Score cannot accurately reflect the performance on adversarial examples (see Appendix~\ref{appdx:biased_IS}), resulting in inconsistency with human evaluation results.}
    \label{fig:ablation}
\end{figure}

We further analyze different components of our method in this section. The analysis is conducted from 3 aspects: 1) the scene prior, 2) the inpainting method and 3) quality enhancement techniques. For each aspect, we remove a component or replace it with other methods, and perform the same evaluation as in the main experiments. We use the same data used in the image-to-panorama main experiment for analysis. The results are reported in Fig.~\ref{fig:ablation}. Please refer to Appendix~\ref{appdx:ablation_vis} for the visualization comparisons.

For the \emph{scene prior}, we remove the prior from chatGPT (no GPT) and all text guidance (no prompt) respectively. Without the global layout prior from chatGPT, the structure of the outputs becomes worse. Without any prompt guidance, both the scene layout and the perspective view rendering quality degrade severely. 

For the \emph{inpainting method}, we replace the Stable Diffusion v2 model with 3 state-of-the-art text-conditioned inpainting methods, namely, Blended Latent Diffusion (BLD inpaint)~\cite{BLD}, Deep Floyd (DF inpaint)~\cite{IF} and Stable Diffusion XL (SDXL inpaint)~\cite{SDXL}. Interestingly, though some of the methods~\cite{SDXL} are published later than Stabld Diffusion v2, their capacity to perform large mask inpainting is limited on in-the-wild images, resulting in worse performance in terms of both scene layouts and rendering quality. Note that the inception scores for some methods (\eg, BLD inpaint) are higher than ours despite a much worse performance from human evaluation. This is caused by the adversarial samples generated by these methods (see Appendix~\ref{appdx:biased_IS} for examples), where the local patches of the adversarial samples are not consistent with the input image, yet leading to a high diversity in the inception score. Similar issues have also been discovered in other problems~\cite{barratt2018note}, which shows the importance of human evaluation.

For the \emph{quality enhancement techniques}, we remove respectively the super-resolution (no SR) and smoothing (no smooth) techniques mentioned in Sec.~\ref{sec:method_quality}. Though the scene layout does not degrade much, the perspective view rendering quality is lower due to the blur or artifacts.


\subsection{Applications}\label{sec:exp_application}

\begin{figure}[t]
    \centering
    \includegraphics[width=1\columnwidth]{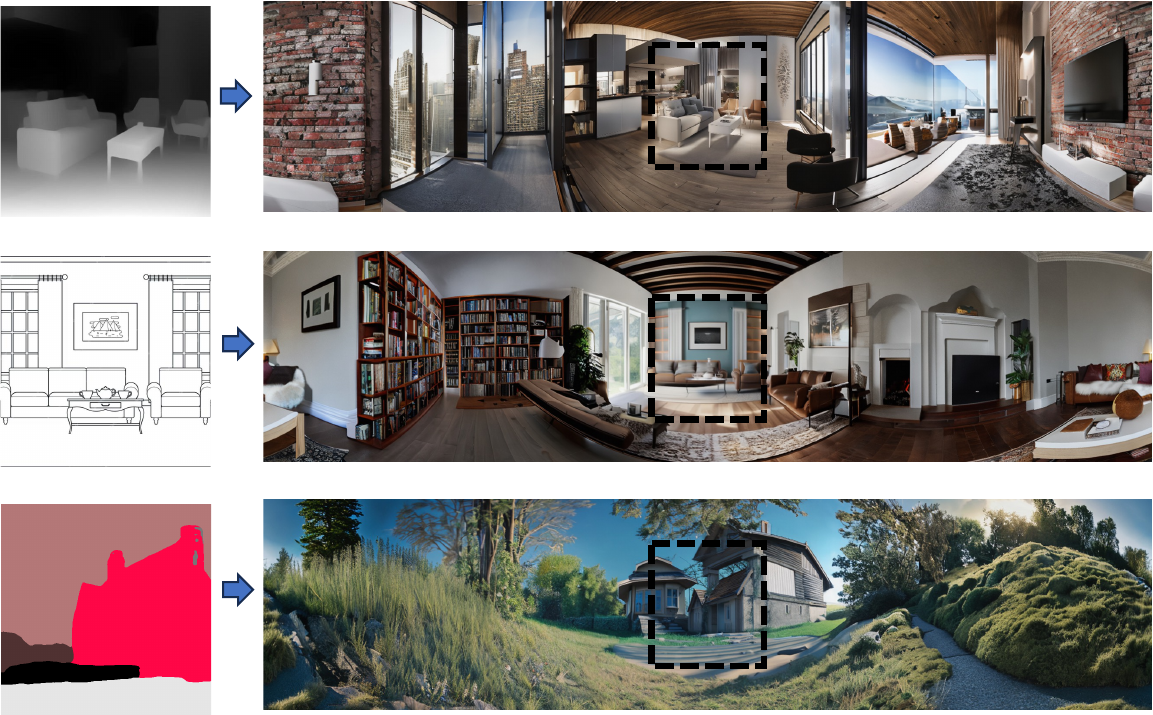}
    \caption{\textbf{Panorama generated from other input modalities.} L-MAGIC can effectively create panoramas from various input modalities, such as an input depth map (top), a sketch drawing (middle) and a colored script or a segmentation mask (bottom). The dotted bounding box indicates the region of the initial perspective view, which is generated by conditional diffusion models.}
    \label{fig:mod}
    \vspace{-1em}
\end{figure}

\begin{figure}[t]
    \centering
    \includegraphics[width=.9\columnwidth]{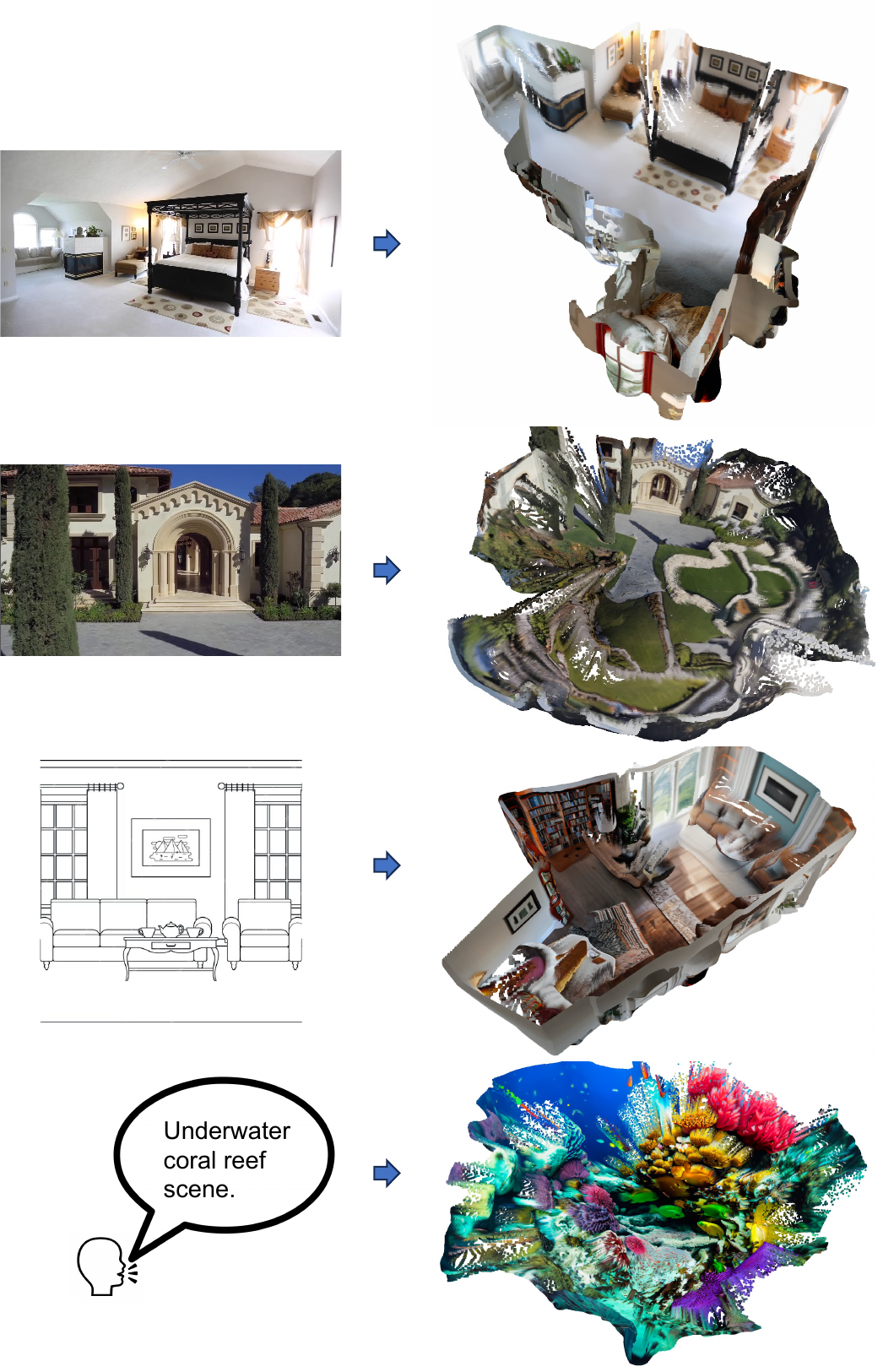}
    \caption{\textbf{3D point cloud generation.} Performing depth estimation on the generated panorama further enables the creation of 3D point clouds from diverse inputs. We can generate point clouds for both indoor and outdoor scenes, even the underwater scene with clear geometry of the fishes and coral reefs.}
    \label{fig:3d}
    \vspace{-1em}
\end{figure}




Combining matured computer vision techniques makes our pipeline applicable to a wide range of applications. 

\mypara{Anything-to-panorama} Conditional diffusion models~\cite{controlnet, stable_diffusion} can now generate an image from diverse types of inputs. The strong zero-shot performance of L-MAGIC makes it possible to generate panoramic scenes from potentially any inputs compatible with conditional diffusion models. As shown in Fig.~\ref{fig:mod}, we can use~\cite{controlnet} to generate a single image from 1) a depth map, 2) a sketch drawing or 3) a colored script or segmentation image. Then, this generated image can be used in L-MAGIC to produce realistic panoramic scenes. This flexibility makes L-MAGIC beneficial to a wide range of design applications.

\mypara{3D scene generation} Applying state-of-the-art depth estimation models, we can further generate 3D scenes from the output of L-MAGIC. Specifically, we compute the depth map for multiple perspective views, then we merge the depth maps into the equirectangular image plane by aligning all views to the initial view. Then we convert the corresponding panoramic depth map to a 3D point cloud. We use Metric3D~\cite{metric3d} and DPT-hybrid~\cite{ranftl2021vision} to estimate the depth for indoor and outdoor scenes respectively. The alignment is done by optimizing the scale and shift of each depth map to enforce the depth from multiple views to be the same at the same pixel. Fig.~\ref{fig:3d} shows sampled results. Despite some artifacts caused by the limitation of monocular depth models, L-MAGIC can generate diverse indoor and outdoor point clouds from various types of inputs. 

\mypara{Immersive video} One can also render immersive videos from our panorama. Specifically, we first generate a panorama using our pipeline, and then generate a series of camera poses for individual video frames. Then we warp the panorama to each frame view according to the camera poses. To further enable scene fly-through with camera translations, we apply depth-based warping when translation is involved in a frame, and inpaint the missing region introduced after translation. See Appendix~\ref{appdx:flythorugh} for implementation details and the supplementary videos for the results.

\section{Conclusion}\label{sec:conclusion}

We have proposed L-MAGIC, a novel method that can generate $360^\circ$ panoramic scenes from a single input image. L-MAGIC leverages large (vision) language models to guide diffusion models to smoothly extend the local scene content with a coherent $360^\circ$ layout. We have also proposed techniques to enhance the quality and resolution of the generated panorama. Extensive experiments demonstrate the effectiveness of L-MAGIC, outperforming state-of-the-art methods for image-to-panorama and text-to-panorama across metrics. Combined with state-of-the-art computer vision techniques such as conditional diffusion models and depth estimation models, our method can consume various types of inputs (text, sketch drawings, depth maps etc.) and generate outputs beyond a single panoramic image (videos with camera translations, 3D point clouds, \etc). See Appendix~\ref{appdx:limit} for discussions about limitations and future works.

{
    \small
    \bibliographystyle{ieeenat_fullname}
    \bibliography{main}
}


\clearpage
\appendix
\section{L-MAGIC Prompts}~\label{appdx:questions}
In Sec.~\ref{sec:inpaint} we have briefly described how to use language models in L-MAGIC. Here, we provide more details on our prompt design when applying language models. 

For line~\ref{line:d_init} of Alg.~\ref{alg:main}, we ask the following two questions to the BLIP-2 model ($\gL_v(\cdot)$):
\begin{enumerate}[label=\textbf{Q\arabic*$_{\text{BLIP}}$}]
    \item \label{Q1} Question: What is this place (describe with fewer than 5 words)? Answer: 
    \item \label{Q2} Question: Describe the foreground and background in detail and separately? Answer: 
\end{enumerate}
These two questions make the model output scene-level coarse and fine descriptions without focusing on centralized objects, which is beneficial for inferring the global scene layout at line~\ref{line:d_360}. The final $d_{\gI}$ is the answers of both questions.

To obtain scene layout descriptions $d_{360}$ of individual views, we ask the following question to ChatGPT ($\gL(\cdot)$) at line~\ref{line:d_360}:
\begin{enumerate}[label=\textbf{Q\arabic*$_{\text{GPT}}$}]
    \item \label{Q1_GPT} Given a scene with $<$answer of~\ref{Q1}$>$, where in font of us we see $<$answer of~\ref{Q2}$>$. Generate 6 rotated views to describe what else you see in this place, where the camera of each view rotates 60 degrees to the right (you dont need to describe the original view, \ie, the first view of the 6 views you need to describe is the view with 60 degree rotation angle). Dont involve redundant details, just describe the content of each view. Also don't repeat the same object in different views. Don't refer to previously generated views. Generate concise ($<$ 10 words) and diverse contents for each view. Each sentence starts with: View xxx(view number, from 1-6): We see...
\end{enumerate}
As mentioned in Sec.~\ref{sec:discussion}, ChatGPT sometimes cannot fully follow the format request in~\ref{Q1_GPT}, which makes automatic prompt generation fail. To avoid this catastrophic failure, we check whether the output of~\ref{Q1_GPT} has the required number of lines (6), and whether each line starts from `\emph{View XXX (line number): We see'}. We re-run line~\ref{line:d_360} if any of the condition is violated. This ensures that ChatGPT understands our question and satisfies all our format requests.

To remove object-level information at line~\ref{line:d_scene}, we ask:
\begin{enumerate}[label=\textbf{Q\arabic*$_{\text{GPT}}$}]
    \setcounter{enumi}{1}
     \item \label{Q2_GPT} Modify the sentence: $<$answer of~\ref{Q1}$>$ so that we remove all the objects from the description (\eg, 'a bedroom with a bed' would become 'a bedroom'. Do not change the sentence if the description is only an object). Just output the modified sentence.
\end{enumerate}

To adaptively judge whether we should avoid repeated objects, we ask the following two questions at line~\ref{line:d_repeat} 
\begin{enumerate}[label=\textbf{Q\arabic*$_{\text{GPT}}$}]
    \setcounter{enumi}{2}
    \item \label{Q3_GPT} Given a scene with $<$answer of~\ref{Q1}$>$, where in font of us we see $<$answer of~\ref{Q2}$>$. What would be the two major foreground objects that we see? Use two lines to describe them where each line is in the format of ``We see: xxx (one object, dont describe details, just one word for the object. Start from the most possible object. Don't mention background objects like things on the wall, ceiling or floor.)''
    \item \label{Q4_GPT} Do we often see multiple $<$each object in the answer of~\ref{Q3_GPT}$>$ in a scene with $<$answer of~\ref{Q1}$>$? Just say 'yes' or 'no' with all lower case letters.
\end{enumerate}
The final $d_\text{repeat}$ is the set of objects in~\ref{Q3_GPT} such that the corresponding answer of~\ref{Q4_GPT} is `no'.


\section{Blur During Warping}\label{appdx:blur}

\begin{figure}[t]
    \centering
    \includegraphics[width=1\columnwidth]{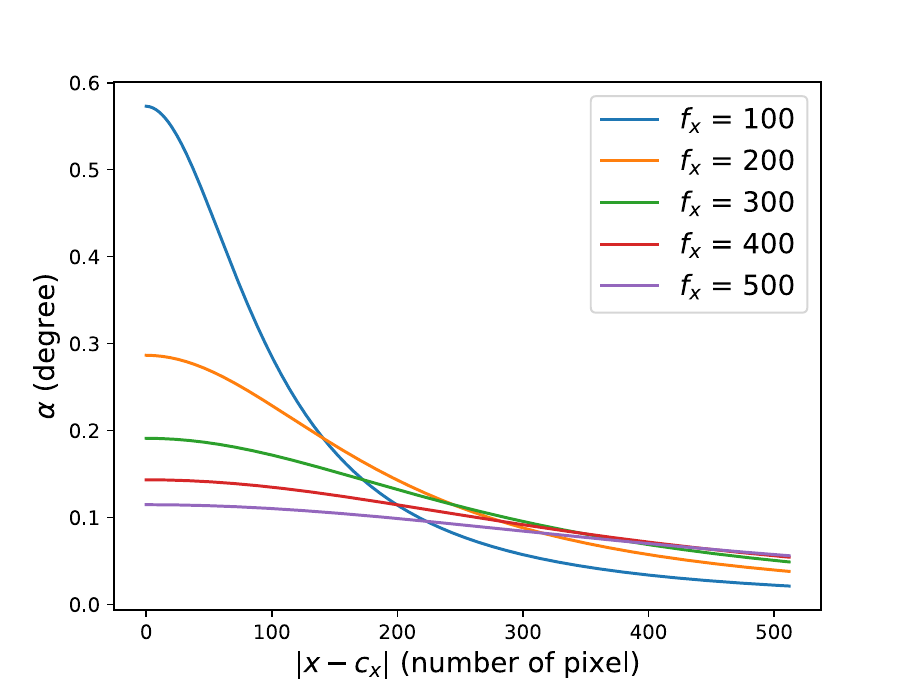}
    \caption{\textbf{Angular distance change w.r.t. the pixel location.} We can see that the angular distance $\alpha$ is larger for centered pixels (small $|x-c_x|$, where $|x-c_x|$ represents the horizontal distance from pixel $x$ to the image center $c_x$) for different focal length values $f_x$. This phenomenon causes the blurry warping mentioned in Sec.~\ref{sec:method_quality}.}
    \label{fig:ang_dis}
\end{figure}

\begin{figure}[t]
    \centering
  \begin{minipage}[b]{0.49\columnwidth}
    \includegraphics[width=\textwidth]{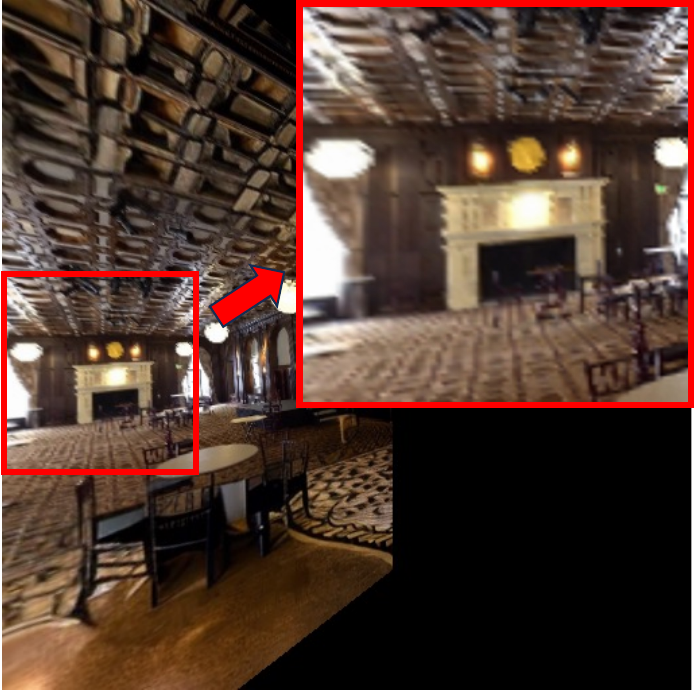}
    \subcaption{w/o super-resolution.}
  \end{minipage}
  \hfill 
  \begin{minipage}[b]{0.49\columnwidth}
    \includegraphics[width=\textwidth]{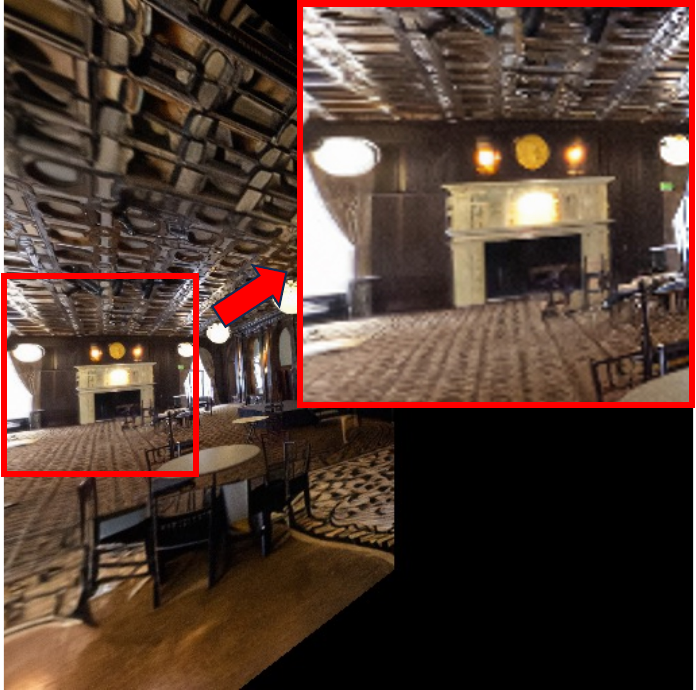}
    \subcaption{With super-resolution}
  \end{minipage}
  \caption{\textbf{Warping with and without super-resolution.} (a) The phenomenon in Fig.~\ref{fig:ang_dis} causes blurry warping. (b) With super-resolution, we can significantly enhance the sharpness of the warped image.}
  \label{fig:warp_blur}
\end{figure}

As mentioned in Sec.~\ref{sec:method_quality}, adjacent pixels at the center of an image have a large angular distance than the ones at the side of an image, which causes the blurry warped image. The cause of this issue lies in the construction process of an image. Specifically, let $x$ be the horizontal coordinates of a pixel at the image plane, and let $f_x$ and $c_x$ be respectively the focal length and the principal location (camera center at the image plane) of the camera on the horizontal direction. Then, the horizontal angular distance between the camera rays of $x_1$ and $x_1 + 1$ is
\begin{equation}
    \alpha = |\arctan(\frac{|x + 1 - c_x|}{f_x}) - \arctan(\frac{|x - c_x|}{f_x})|
\end{equation}
Fig.~\ref{fig:ang_dis} shows the change of value $\alpha$ w.r.t. $|x-c_x|$, where large $|x-c_x|$ means $x$ is at the side of an image, and small $|x-c_x|$ means x is at the center of an image. We can see that the angular distance $\alpha$ is larger for centered pixels regardless of the focal length $f_x$. Hence, within the same angle, there are more pixels on the side of an image than at the center of an image. This means that when warping the center region of an image to another view, we require interpolation since more pixels are created in the corresponding warped region. This phenomenon causes the blurry warping mentioned in Sec.~\ref{sec:method_quality}, see Fig.~\ref{fig:warp_blur} for an example.

\section{Text Inputs for Text-to-panorama}\label{appdx:text2pano}
In order to evaluate the performance of different algorithms on in-the-wild inputs, we ask ChatGPT to generate 20 random scene descriptions (10 indoor and 10 outdoor) in the main experiment of text-to-panorama (Sec.~\ref{sec:exp_main}). The resulting text prompts are:
\begin{enumerate}
    \item Autumn maple forest path.
    \item Tropical beach at sunset.
    \item Snowy mountain peak view.
    \item Tuscan vineyard in summer.
    \item Desert under starlit sky.
    \item Sakura blossom park, Kyoto.
    \item Rustic Provencal lavender fields.
    \item Underwater coral reef scene.
    \item Ancient Mayan jungle ruins.
    \item Manhattan skyline at night.
    \item Victorian-era library.
    \item Rustic Italian kitchen.
    \item Minimalist Scandinavian bedroom.
    \item Moorish-styled bathroom.
    \item Vintage record store interior.
    \item Luxurious Hollywood dressing room.
    \item Industrial loft-style office.
    \item Art Deco hotel lobby.
    \item Japanese Zen meditation room.
    \item Modern living room with a sofa and a TV.
\end{enumerate}

\section{Voting Web Page}\label{appdx:eval_webpage}
\begin{figure}[t]
    \centering
    \begin{minipage}[b]{0.49\textwidth}
    \includegraphics[width=\textwidth]{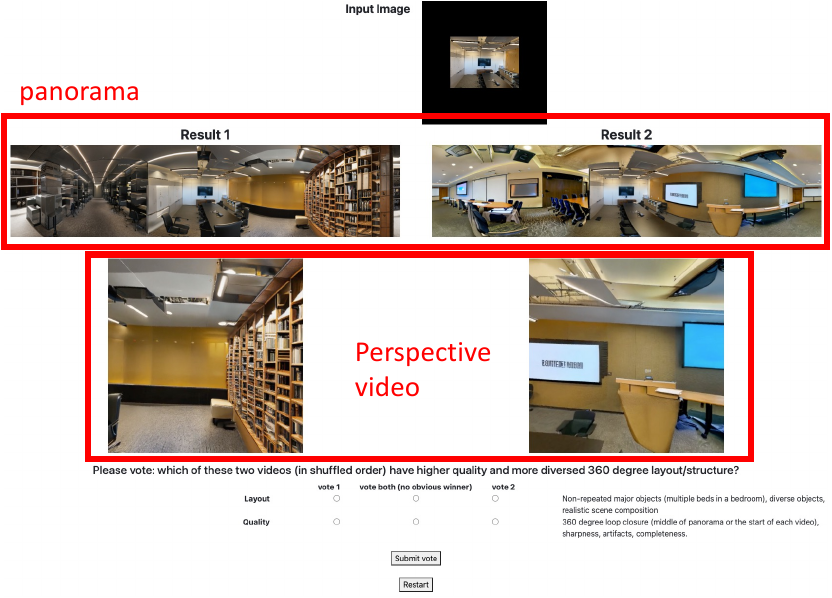}
    \subcaption{Image-to-panorama.}
    \end{minipage}
    \begin{minipage}[b]{0.49\textwidth}
    \includegraphics[width=\textwidth]{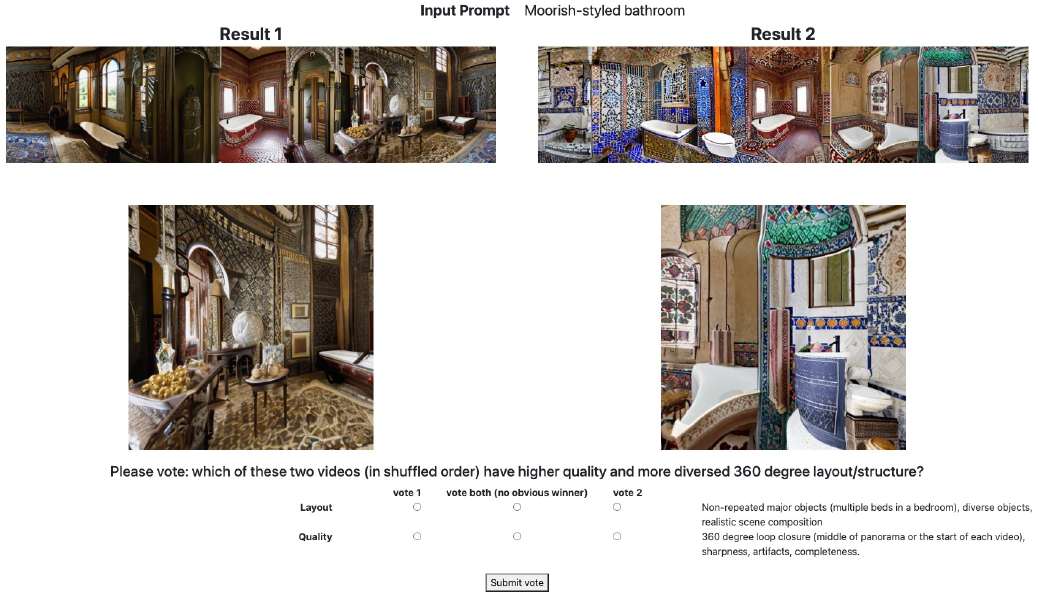}
    \subcaption{Text-to-panorama}
    \end{minipage}
    \caption{\textbf{The voting web page for human evaluations.} (a) The web page for image-to-panorama. The outer black region of the input image is the missing region when expanding the field of view. (b) The web page for text-to-panorama. In each voting, we show the panoramas and the rendered perspective videos for two methods. For each criterion, we not only allow to vote for one of the results but also allow to vote for both results when there is no obvious winner.}
    \label{fig:vote}
\end{figure}
As mentioned in Sec.~\ref{sec:exp_setup}, we use a voting web page during human evaluation. Fig.~\ref{fig:vote} shows the example web page for both image-to-panorama and text-to-panorama. In each voting, we show for each method a panorama and the perspective video rendered from the panorama so that the user can use the panorama to clearly see the 360 degree layout and loop closure, and use the perspective video to see the rendering quality. Besides voting for one of the two results, we also allow to vote for both results when there is no obvious winner for a certain criterion.

\section{Ablation visualizations}\label{appdx:ablation_vis}

\begin{figure*}[t]
    \includegraphics[width=1\textwidth]{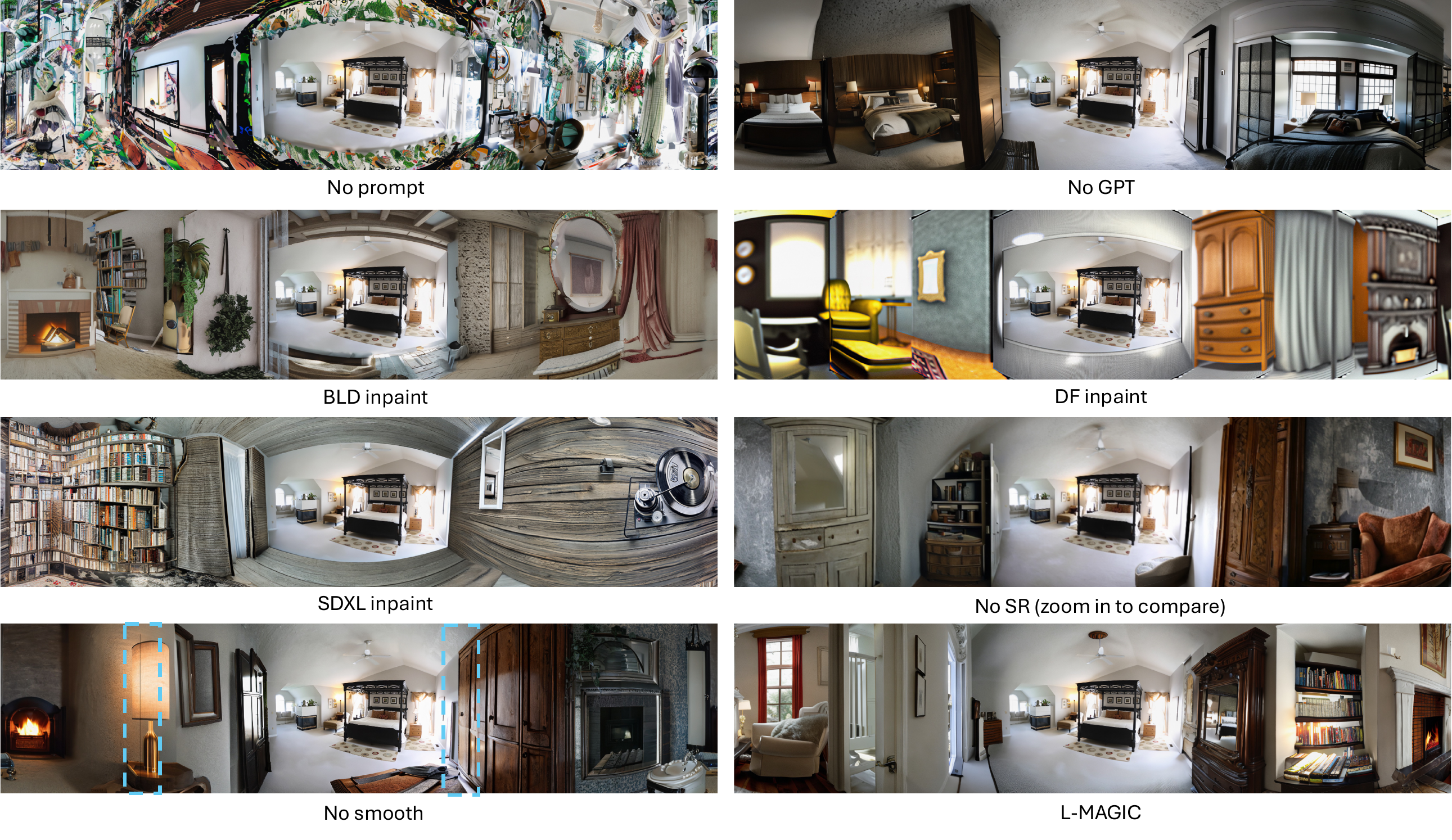}
\caption{\textbf{Visualization of ablation results.} Consistent with the quantitative result in Sec.~\ref{sec:exp_analysis}, removing individual components of L-MAGIC hurt the performance. A zoom-in comparison is recommended for \emph{No SR} and \emph{No smooth}. In \emph{No SR}, the panorama is less sharp even though the image resolution is the same with L-MAGIC. In \emph{No smooth}, there were two unnatural black lines (zoom in to the bounding boxes) caused by the non-smooth fusion, which we do not observe in the full L-MAGIC method.}\label{fig:vis_ablation}
\end{figure*}

In Sec.~\ref{sec:exp_analysis}, we conducted ablation studies and reported quantitative results. Here we further show visualizations of the ablation experiment in Fig.~\ref{fig:vis_ablation}. Consistent with the quantitative results, changing L-MAGIC components hurts the visual quality of the output panorama.

\section{Bias in Quantitative Metrics}\label{appdx:biased_IS}
\begin{figure*}[t]
\includegraphics[width=1\textwidth]{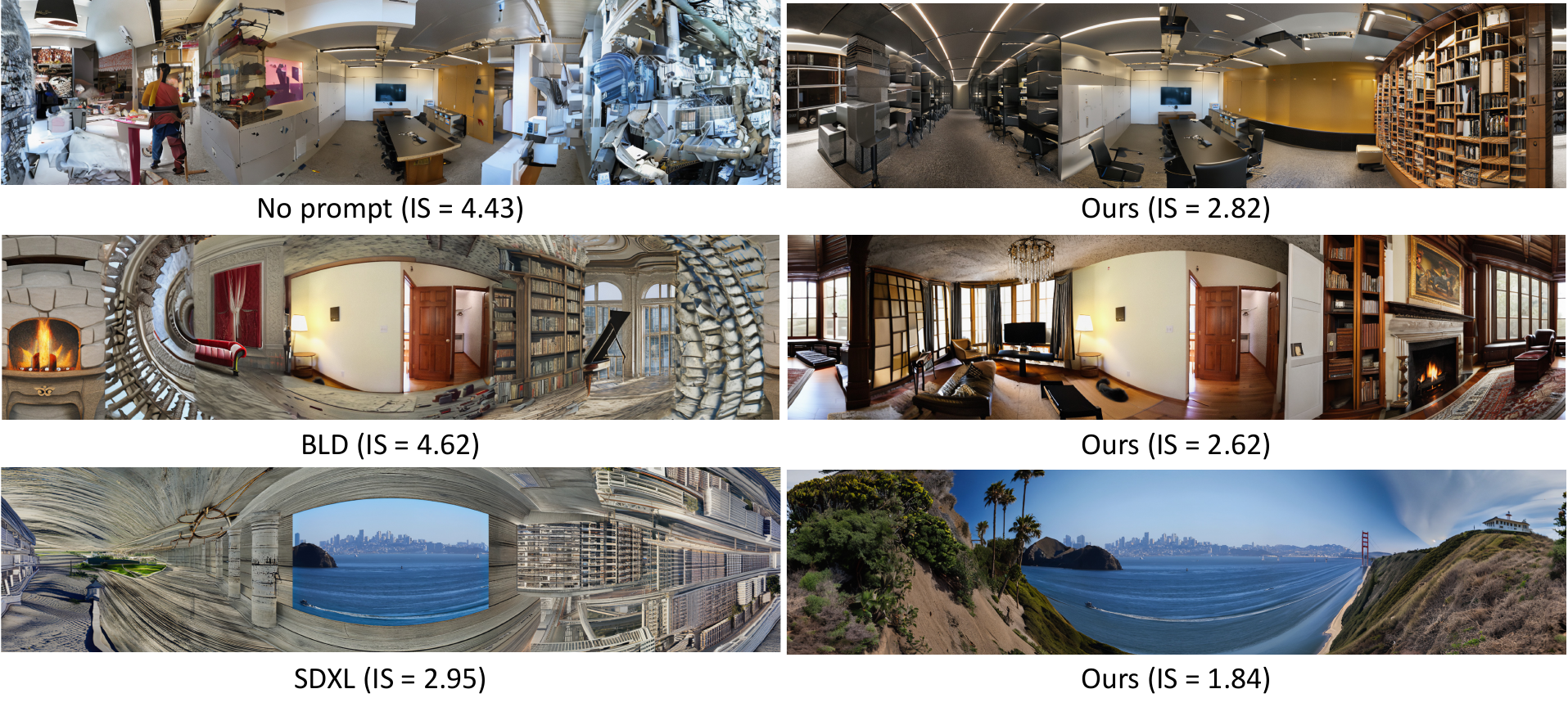}
\caption{\textbf{Example of the biased Inception Score.} We show samples with high inception scores from the experiment in Fig.~\ref{fig:ablation} (Left). Even though our method (right) generates scenes with a much better quality and multi-view consistency, the inception score can still be lower.}\label{fig:bias_IS}
\end{figure*}
As mentioned in Sec.~\ref{sec:exp_analysis}, the Inception Score (IS) sometimes cannot fully reflect the preference from human evaluations. Fig.~\ref{fig:bias_IS} shows "adversarial" examples where the panorama has poor quality and multi-view coherence yet has a higher inception score compared to the result with better human evaluation preference. This shows the importance of human evaluations in the experiment. 

\section{Video generation}\label{appdx:flythorugh}

\begin{figure}[t]
    \centering
    \begin{minipage}[t]{0.49\columnwidth}
        \includegraphics[width=1\textwidth]{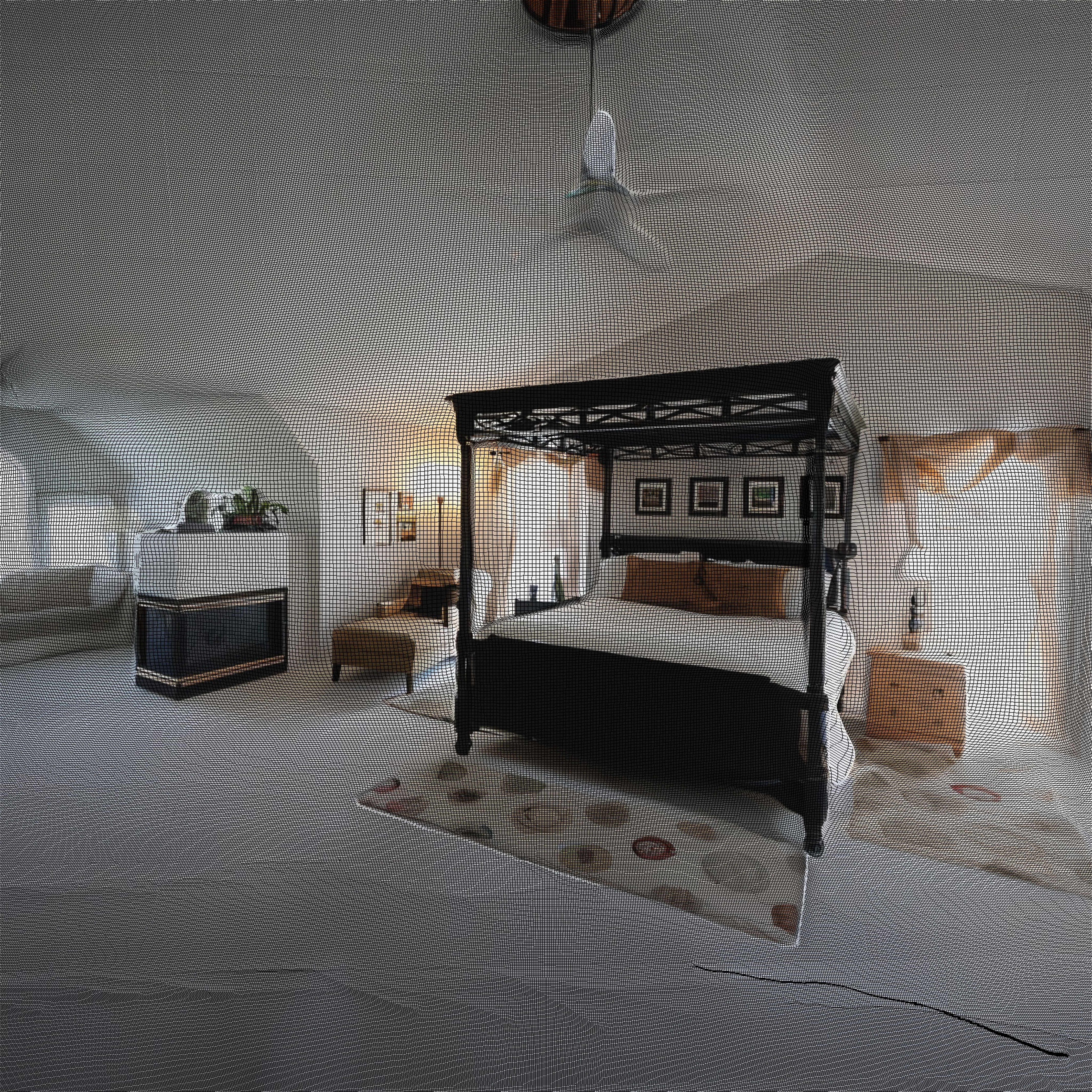}   
    \end{minipage}
    \begin{minipage}[t]{0.49\columnwidth}
        \includegraphics[width=1\textwidth]{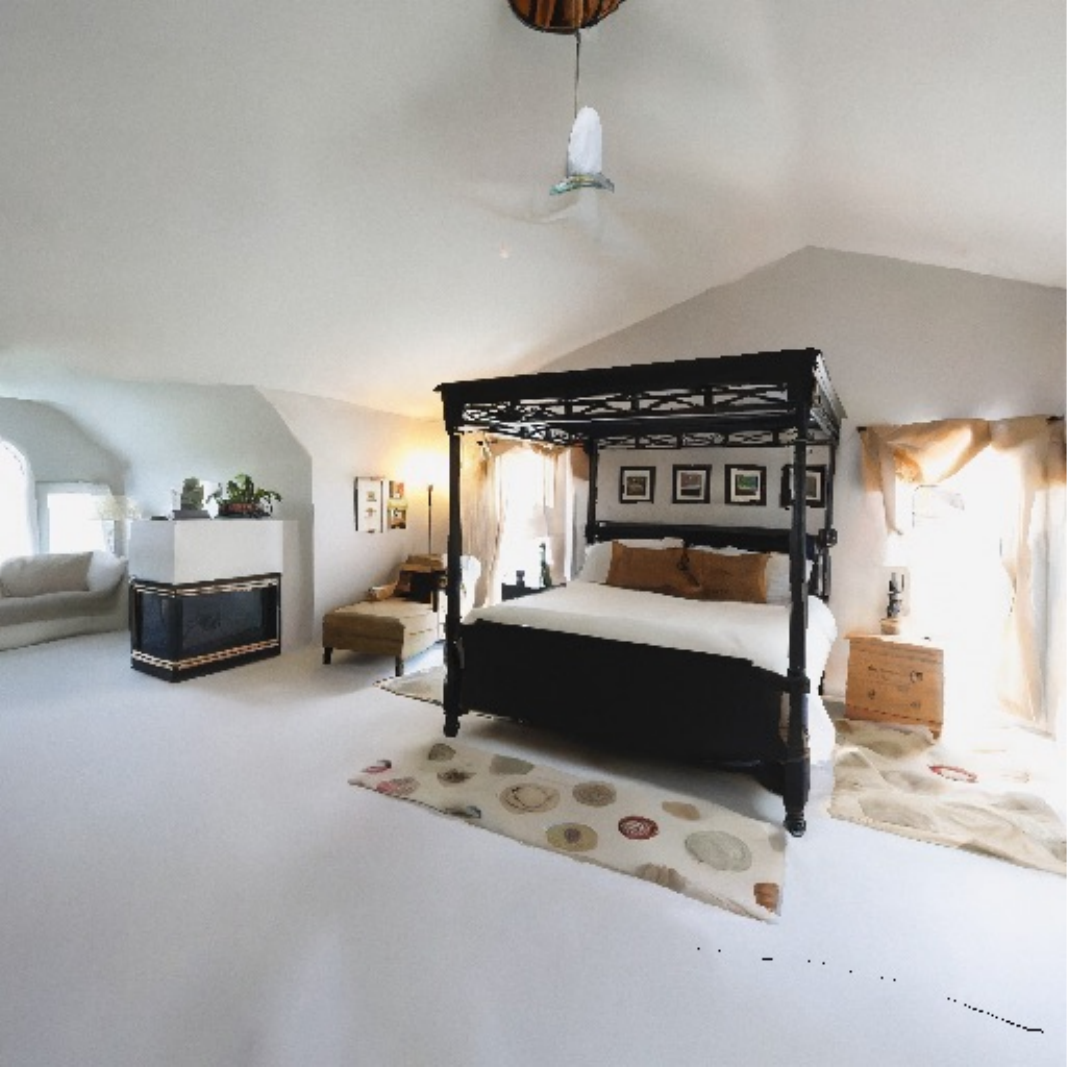}           
    \end{minipage}
    \caption{\textbf{Effect of super-resolution on depth-based warping.} Left: Naive point-based warping generates grid-shaped missing pixels distributed uniformly on the whole image. Right: Super-resolution effectively resolves this issue, making most parts of the warped image sharp and complete.}
    \label{fig:super-depth}
\end{figure}

When generating video frames with pure camera rotation, we follow the strategy of Sec.~\ref{sec:warp}, project the panorama to a unit sphere, and render each frame according to the rotation matrix and camera intrinsics of the frame.

To generate immersive videos with camera translations, we apply depth-based warping, and inpaint, using Stable Diffusion v2, the small missing regions caused by occlusion. For depth-based warping, we first apply pre-trained depth estimation models~\cite{ranftl2021vision} on perspective views of the generated panorama, and warp them to the corresponding frame of the video. Naive mesh-based warping following Sec.~\ref{sec:warp} may generate mesh faces between different objects, which is not ideal. Hence, we rely on point-based warping. To avoid grid-shaped sparsely distributed missing pixels (Fig.~\ref{fig:super-depth} left) and ensure the sharpness of the warped image, we apply a super-resolution-based approach similar to the strategy in Sec.~\ref{sec:method_quality}. Specifically, we enlarge the resolution of the depth map from $512 * 512$ to $2048 * 2048$, and then warp the high-resolution depth map to each frame with a resolution of $512 * 512$ (Fig.~\ref{fig:super-depth} right). 

To achieve super-resolution on the depth map, we first perform super-resolution on the RGB image, increasing its resolution to $2048 * 2048$. Since state-of-the-art depth estimation models are not effective on high-resolution images, instead of directly estimating the depth of the high-resolution image, we separate it into $13 * 13$ patches of resolution $512 * 512$ with overlappings between neighbouring patches, perform depth estimation on individual patches and align the depth map of each patch with the one of the low-resolution image to ensure a smooth depth transition over patches and a reasonable object geometry.

\section{Limitation and Future Work}\label{appdx:limit}

In terms of the limitation, L-MAGIC currently relies on the input prompt to encode the global scene layout information. Designing a fine-grained layout encoding mechanism that can ensure multi-view coherence at a more detailed level is an important and interesting future work. 

\end{document}